\def\year{2022}\relax
\documentclass[letterpaper]{article} 
\usepackage{aaai22}  
\usepackage{times}  
\usepackage{helvet}  
\usepackage{courier}  
\usepackage[hyphens]{url}  
\usepackage{graphicx} 
\usepackage{natbib}  
\usepackage{caption} 
\DeclareCaptionStyle{ruled}{labelfont=normalfont,labelsep=colon,strut=off} 
\usepackage{algorithm}
\usepackage{algorithmic}

\usepackage{soul}
\usepackage{xcolor}
\usepackage{latexsym}
\usepackage{comment}

\usepackage{microtype}
\usepackage{newfloat}
\usepackage{listings}
\floatstyle{ruled}
\newfloat{listing}{tb}{lst}{}
\floatname{listing}{Listing}
\newcommand{\fbpds}[0]{fpb75}
\newcommand{\yelpds}[0]{yelp}
\newcommand{\sstds}[0]{sst2}
\newcommand{\amazonds}[0]{amazon}

\newcommand{\gencorpus}[0]{$C_{g}$}
\newcommand{\fincorpus}[0]{Finance Corpus}

\newcommand{\scicorpus}[0]{Science Corpus}

\newcommand{\sportscorpus}[0]{Sports Corpus}

\newcommand{\genmodel}[0]{$SenDM$}
\newcommand{\genmodelnonfin}[0]{$SenDM^{\ast}$}
\newcommand{\genmodelnonfinbold}[0]{$\bf{SenDM^{\ast}}$}
\newcommand{\genmodelbase}[0]{SenDM-base}
\newcommand{\genmodeltiny}[0]{SenDM-tiny}
\newcommand{\dmodelp}[0]{$SenDM^{P}_d$}
\newcommand{\dmodelLg}[0]{$SenDM^{L_g}_d$}
\newcommand{\dmodelLgP}[0]{$SenDM^{L_g+P}_d$}
\newcommand{\dmodelLd}[0]{$SemDM^{L_d}_d$}
\newcommand{\dmodelLdP}[0]{$SemDM^{L_d+P}_d$}
\newcommand{\berttiny}[0]{BERT-tiny}
\newcommand{\bertbase}[0]{BERT-base}

\newcommand{\finbert}[0]{FinBERT}
\newcommand{\sentix}[0]{SentiX}

\title{Fortunately, Discourse Markers Can Enhance Language Models for Sentiment Analysis}

\author{Liat Ein-Dor\thanks{\ \ These authors equally contributed to this work.}\textsuperscript{\rm 1}, Ilya Shnayderman$^{*}$\textsuperscript{\rm 1}, Artem Spector$^{*}$\textsuperscript{\rm 1}, Lena Dankin\textsuperscript{\rm 1},\\ Ranit Aharonov\thanks{Current affiliation:  Pangea Therapeutic, ranitah1@gmail.com}\textsuperscript{\rm 1} and Noam Slonim\textsuperscript{\rm 1}}

\affiliations{
    \textsuperscript{\rm 1}IBM Research\\
    \{liate,ilyashn,artems,lenad,noams\}@il.ibm.com\\

}

\usepackage{bibentry}

\begin{document}

\maketitle
\begin{abstract}
\begin{quote}
In recent years, pretrained language models 
have revolutionized the NLP world, while achieving state-of-the-art performance in various downstream tasks. 
However, in many cases, these models do not perform well when labeled data is scarce and the model is expected to perform in the zero or few shot setting.
Recently, several works have shown that 
continual 
pretraining or 
performing 
a second phase of pretraining (inter-training), which is better aligned with the downstream task, can lead to improved results, especially in the scarce data setting.
Here, we 
propose to leverage 
sentiment-carrying 
discourse-markers 
to generate 
large-scale weakly-labeled data, which in turn can be used to adapt general-purpose language models to the task of 
sentiment classification. 
In addition, we propose a new method for adapting sentiment classification models to new domains, which is based on automatic identification of domain-specific sentiment-carrying 
discourse markers.
Extensive experimental results show the value of our approach on various benchmark datasets. 
Code, models and data
are available at \url{https://github.com/ibm/tslm-discourse-markers}.



\end{quote}
\end{abstract}

\section{Introduction}

Large pretrained language models are reshaping the landscape of NLP. These models, recently referred to as {\it foundation models\/} \cite{bommasani2021opportunities}, were originally proposed with a two-step paradigm in mind. The model is first pretrained at scale on broad data with a surrogate self-supervised task; the knowledge gained by this pretraining is then transferred and adapted via fine-tuning on -- typically small -- labeled data, to a specific downstream task. Prominent examples include BERT \cite{bert} and GPT-3 \cite{gpt-3}. The practical value of this approach is immense. The self-supervised pretraining requires no labeled data. The resulting model represents a single powerful starting point that can be swiftly adapted to address a wide range of target tasks with relatively little annotation effort, via few-shot or even zero-shot learning \cite{gpt-3}. 

Subsequent studies have shown that the original two-step paradigm can be further refined to yield an even better starting point model for particular tasks of interest.
For example, continuing the pretraining on domain-specific data such as finance or legal documents have proven beneficial to tasks in these domains 
\cite{Araci2019FinBERTFS,chalkidis-etal-2020-legal,Donotstoppretraining}.
Similarly, additional pretraining of BERT on dialog data yields better results in target tasks related to dialogue application \cite{wu-etal-2020-tod}, and continual pretraining of BERT on product reviews with sentiment-aware pretraining tasks led to improved performance in sentiment analysis in this domain \cite{zhou-etal-2020-sentix}. Another, more computationally demanding option is to pretrain the model from scratch on self-supervised task(s) that aim to better reflect the nature of the target tasks. For example, the pretraining tasks of SpanBERT \cite{DBLP:journals/corr/abs-1907-10529} and PEGASUS \cite{pmlr-v119-zhang20ae} are designed to be closer in spirit to span-extraction tasks as in question answering and to summarization tasks, respectively, resulting in better performance in these target tasks.

A related path, which is further explored in this work, is to add an intermediate training step, referred to as {\it inter-training\/}, which is somewhat aligned with a specific target task of interest. 
There are several 
aspects by which these 
inter-training 
approaches differ.
One main aspect is the similarity between the intermediate task and the target task which ranges from full alignment using weakly 
or readily available 
labeled data \cite{meng2020text,zhou-etal-2020-sentix,huber2021ccqa} to transfer learning using labeled data on a similar yet 
different task \cite{pruksachatkun2020intermediate}, and 
further including 
works which perform transfer learning with no labeled data, e.g., \cite{shnarch2021cluster} apply 
unsupervised text clustering and then inter-train 
a model to predict the cluster label.
Among the approaches that rely on fully aligned intermediate tasks, some works leverage weak labels that are inherent to the original 
text, like the explicit mention of the class name 
\cite{meng2020text} or 
the presence of the token 'that' in a sentence 
\cite{levy2018towards}; 
while others rely on non-textual signals like human-added numeric review ratings 
\cite{zhou-etal-2020-sentix} or sentiment-bearing emojis 
\cite{8560896}. 
Weak labels that are inherent to the text 
usually have limited coverage and involve bias towards specific keywords or patterns 
that define 
the weak signal. 
While the non-textual signals usually do not suffer from these issues, since they are external to the text, they are often specific to task and domain and therefore 
are less directly applicable to new tasks and/or in new domains. 

The present work suggests a new type of weak labels which are 
inherent to the original text, 
but at the same time 
can be perceived as an external label that can be removed from the original text while keeping
the remaining text 
meaningful and grammatical \cite{moder2004discourse}.

Specifically, we propose to leverage the signal carried by particular discourse markers (DMs) to generate large amounts of weakly labeled data for the important task of sentiment analysis (SA).
For example, we assume that sentences following the prefixes "Happily," and "Sadly," convey a positive sentiment and a negative sentiment, respectively. Exploiting this simple assumption with a small seed of $11$ discourse markers, we generate large amounts of weakly labeled data out of a large and general English corpus. Inter-training \bertbase{} and \berttiny{} on this data yields significant performance improvements, especially when labeled data is scarce and in a zero-shot scenario. Moreover, we show how to use the obtained classifier to automatically reveal sentiment-carrying discourse markers in particular domains. Relying on these domain-specific sentiment-carrying discourse-markers yields an additional performance 
gain in zero-shot learning, 
and may further open the door for additional future applications. 
In summary, our main contributions are:
\begin{enumerate}
    \item A novel approach that leverages sentiment signals of discourse markers for creating sentiment-aware language models that significantly outperform prior models. 
    \item A new method for enhancing domain-specific sentiment classification, based on statistical analysis of discourse markers in a domain-specific corpus.
    \item A large dataset of weakly labeled sentences from Wikipedia, and a code for generating weakly labeled data from a given text corpus.
\end{enumerate}

\section{Related Work}
\textbf{Learning with Discourse Markers} Discourse markers (DMs) are words or phrases 
that play a role in managing the flow and structure of discourse. 
DMs have been used as a learning signal for the prediction of implicit discourse relations \cite{liu-li-2016-recognizing,braud-denis-2016-learning} and inference relations \cite{pan-etal-2018-discourse}. 
The task of DM prediction has been leveraged in several works such as \cite{jernite2017discoursebased,nie-etal-2019-dissent,sileo-etal-2019-mining}, to learn general representations of sentences, which can be transferred 
to various NLP classification tasks. \citet{sileo2020discsense} were the first to systematically study the association between \textit{individual} DMs and \textit{specific} downstream task classes. Using a model trained to predict discourse markers between sentence pairs, they predict plausible markers between sentence pairs with a known semantic relation (provided by existing classification datasets). Based on these predictions, they study the link between discourse markers and the semantic relations annotated in classification datasets.
 Here we show 
how such an 
association can be leveraged to enhance the performance of language models on a downstream task, 
and furthermore in a particular domain. 

\textbf{Task-aware Language models.}
A recent line of works has been focused on bridging the gap between the self-supervision task and the downstream tasks which is inherent to multi-purpose pretrained models \cite{sun2019ernie,tian2020skep,chang2020pre}. 
In \citet{joshi2020spanbert}, spans of texts are masked rather than single tokens, resulting in a language model oriented to span-selection tasks. 
\citet{chang2020pre} suggested a language model targeted at document retrieval,
and 
\citet{zhang2020pegasus} pursued a similar goal 
for abstractive text summarization. 
For sentiment analysis, several works have incorporated sentiment knowledge into the pretraining task \cite{tian2020skep,gu2020train}, while focusing mainly on word-level sentiment prediction tasks. Here, in order to achieve full alignment with the downstream task of sentence-level sentiment classification, we suggest a model that incorporates a \textit{sentence-level} sentiment prediction objective. Similar objective was used in \citet{zhou-etal-2020-sentix}, 
relying 
on ratings as sentiment signals, which are specific to the reviews domain. 
In contrast, 
our approach relies on sentiment signals that are carried by discourse markers, which are an inherent to language itself and are 
therefore available for a wide range of domains.

\section{\genmodel{}: A New Sentiment Language Model}

\subsection{Training DM-based Sentiment Models}
\label{section:the_alg}

\begin{figure*}
\centering\includegraphics[scale=0.3]{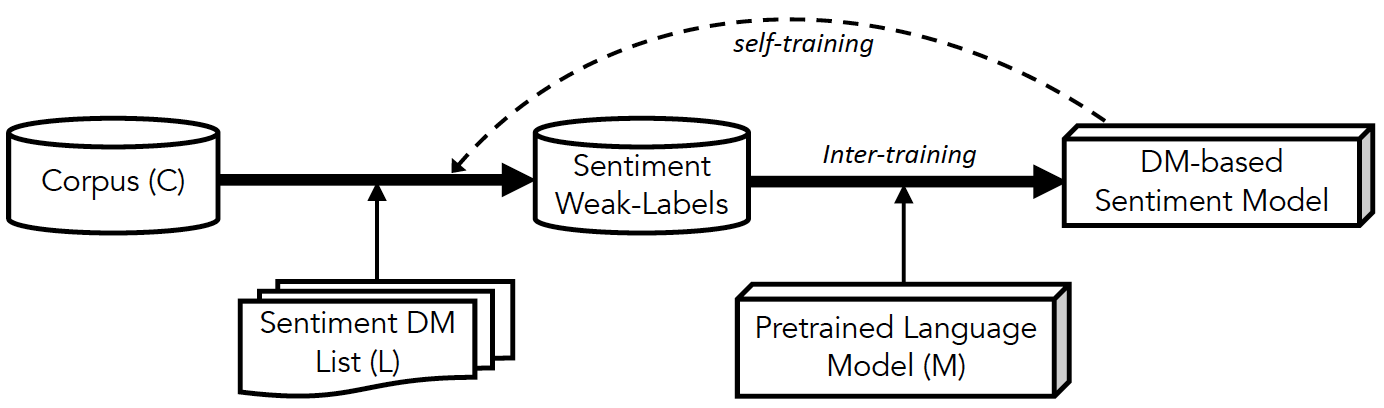}

\caption{Overview of how DM-based sentiment models are trained.}
\label{fig:sketch}
\end{figure*}

We propose a general approach to develop DM-based sentiment models. Our approach relies on weakly labeled sentiment data set, which is automatically 
derived from a given corpus 
by leveraging strong associations between 
DMs 
and sentiment classes, as depicted 
in figure \ref{fig:sketch}.
Given a corpus $C$ and a list $L$ of DMs that signal either a positive or a negative sentiment, each accompanied by its class label, we follow the heuristic introduced by \citet{rutherford2015improving}
and look for all sentences in $C$ that start with $l \in L$ followed by a comma.
We then remove $l$ and the comma from the beginning of each sentence, and annotate all resultant sentences with the class label associated with $l$. 
This process results in a binary classification dataset for sentiment analysis, which is used to fine-tune a pre-trained language model, $M$ (\textit{inter-training}).
In this work we use the above flow to generate a new sentiment model, \genmodel{}, and also to build
an 
additional domain-adapted model 
as will be discussed in section \ref{sec:finance-model}. 

\subsection{The \genmodel{} model} 
We introduce \genmodel{}, a general sentiment model, that aims to improve 
the 
performance of sentiment classification across domains. \genmodel{} is obtained using the flow described above, where $C$ is a general corpus of newspaper and journal articles, denoted \gencorpus{} (see section \ref{sec:gen-exp-setup}), 
and $L$ is a seed 
list of sentiment related DMs obtained manually using general knowledge of the English language. 
More specifically, we asked $3$ annotators, to go over a list of $173$ commonly used 
DMs described in \citet{sileo-etal-2019-mining}, and mark any DM as positive/negative if it is likely to open 
a sentence bearing a positive/negative sentiment, based on its common usage in the English language. The final list, $L_{g}$, consists of $11$ DMs, selected by all $3$ annotators. The DMs identified as associated with a positive sentiment are: \textit{'luckily', 'hopefully', 'fortunately', 'ideally', 'happily', and 'thankfully'}. Those associated with a negative sentiment are: \textit{'sadly', 'inevitably', 'unfortunately', 'admittedly', and 'curiously'}.
The resulting weakly labeled data is used to 
fine tune both the uncased base and tiny architectures of BERT 
\cite{bert,jiao-etal-2020-tinybert}; 
We 
denote the resulting models by \genmodelbase{} and \genmodeltiny{}, respectively,
and 
release both of these models as part of this work. 

\subsection{Experimental Setup}
\label{sec:gen-exp-setup}

\subsubsection{The General Corpus (\gencorpus{})} 
Our proposed solution relies on 
the availability of 
a corpus of unlabeled text. We use a corpus of some $400$ million newspaper and journal articles\footnote{From the LexisNexis 2011-2018 corpus,https://www.lexisnexis.com/en-us/home.page}, breaking the articles into sentences, and indexing these sentences.
We focus on English sentences
\footnote{Specifically, sentences with probability $> 75\%$ of being English, based on Fast-Text langid from \cite{grave-etal-2018-learning}.} and following \citet{sileo-etal-2019-mining} we use only sentences which are $3-32$ tokens in length and have balanced parentheses.

\subsubsection{Inter-training Details.}
The inter-training step (Figure \ref{fig:sketch}) consists of fine-tuning BERT using weakly labeled data.
For inter-training \genmodel{}, we obtain a total of $1,876,614$ weakly labeled sentences, by using the list of sentiment-related DMs, $L_g$, over sentences in 
$C_g$, as described in section \ref{section:the_alg}. 
We divide the samples into training ($80\%$), development ($10\%$), and test ($10\%$) sets. We set the learning rate to $5e-5$, and the batch size to $32$. We use the early stopping strategy, setting the max number of epochs to $4$ and selecting the model with the best accuracy on the development set. The dropout probability is always kept at $0.1$. We employ an Adam optimizer with $\beta_{1} = 0.9$, $\beta_{2} = 0.999$, and $\epsilon = 1e-06$. 
 Training is performed on two V100 GPUs. 

\subsection{Evaluation Details}
\label{sec:eval}
Evaluation is performed on the  datasets appearing below, in three scenarios: zero-shot, few-shot and full-data. 
For zero shot we simply use the classification layer obtained from inter-training.
For few-shot, we further fine-tune the inter-trained model with a small sample of $n$ examples from the training set, with $n$ ranging from $16$ to $1024$. We repeat each experiment five times with different random seeds, each time selecting different examples for fine-tuning.  
In the full-data scenario, all training examples are used for fine-tuning.

To support training on small samples,  the batch size is set to $16$. The other hyper-parameters are the same as in the inter-training phase described above, with one exception. For the few-shot scenario, which represents a low resource setting, we assume that no development set is available for employing the early stopping strategy. Instead, we follow the observation in \citet{zhang2020revisiting} that for small training data, more iterations help stabilize BERT results, and set the number of epochs to $10$. 

\subsubsection{Datasets}

The datasets used for evaluation are presented in Table \ref{tab:datasets}. All datasets contain sentences that are labeled for sentiment. 
\amazonds, \sstds, and \yelpds{} consist of review sentences. \fbpds{} is comprised of sentences from financial news.
 Most of these datasets provide more than two possible labels, so we adjust the datasets for the binary sentiment classification task. 
Specifically, \fbpds{} contains sentences that are labeled as neutral, which we remove from the training and test sets. 
\amazonds{}  and \yelpds{} contain five different labels that reflect 
the sentiment ratings of each sentence ("stars"). We leave only sentences with the lowest and highest scores, considering those as negative and positive labels,
respectively. 
For fine tuning we use up to $1024$ examples from the training set.
For testing we use the entire test set.

\begin{table}[ht!]
\begin{center}
\begin{tabular}{l| l |l} 
 \hline
 \textbf{Dataset} & 
 \textbf{Domain} &
  \textbf{Test set size} \\ 
 \hline
 \amazonds{} & Product reviews & 2K \\
 \hline
\yelpds{}  & Business reviews &  20K \\ 
\hline
\sstds{} & Movie reviews &  1821 \\ 
\hline
\fbpds{}  & Financial news &  691\\
\hline
\end{tabular}
\end{center}
\caption{Datasets used for evaluation. References for the datasets are as follows, by the order appearing in the table: \protect\cite{marc_reviews}, \protect\cite{yelp_dataset}, \protect\cite{wang-etal-2018-glue}, \protect\cite{Malo2014GoodDO}.}
\label{tab:datasets}
\end{table}

\subsection{Results}
\label{sec:resgenmodel}

\begingroup
\setlength{\tabcolsep}{1pt} 
\renewcommand{\arraystretch}{0.2} 

\begin{figure*}
\begin{tabular}{ccc}
    \begin{tabular}{c}
    \\   \raisebox{30ex}{ \amazonds} \\ \raisebox{31ex }{\yelpds}
    \\  \raisebox{1ex }\sstds \\   \raisebox{-30ex }\fbpds
 \end{tabular} &
 
 \begin{tabular}{c}
    base \\ \includegraphics[width = 2.64in]{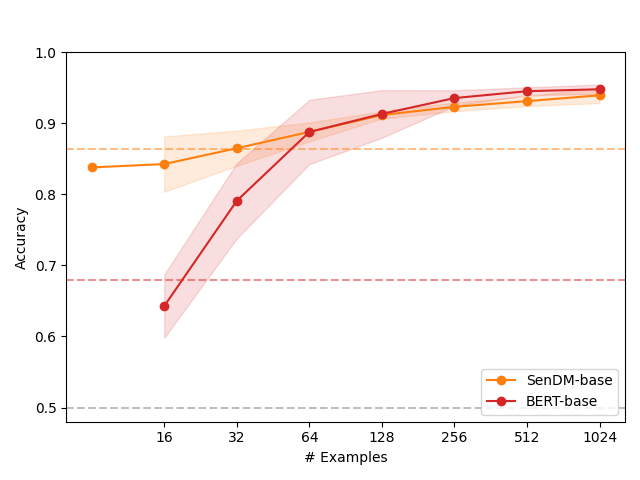} \\
    \includegraphics[width = 2.64in]{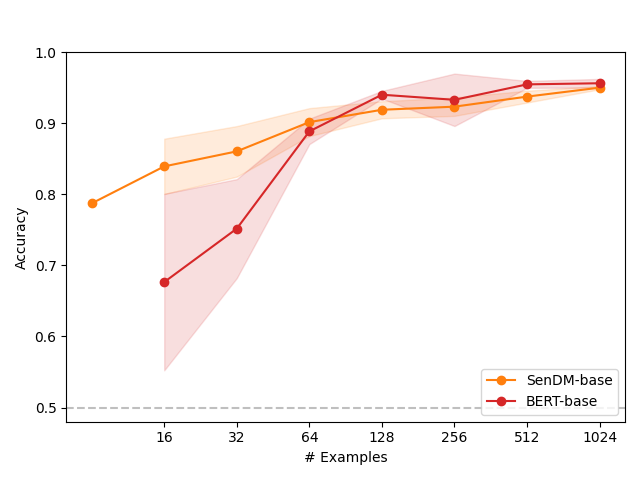}
    \\ \includegraphics[width = 2.64in]{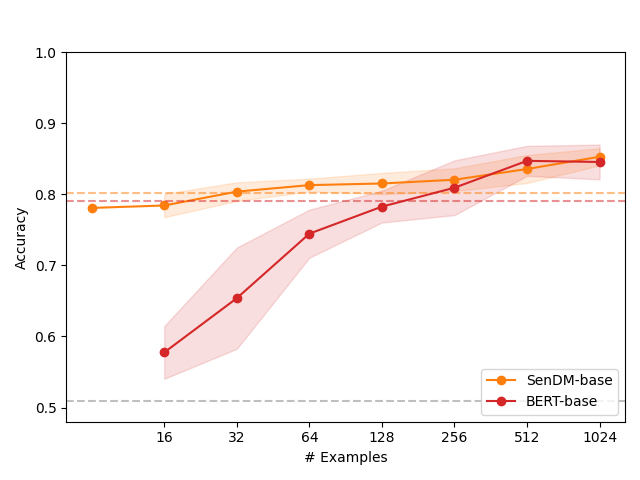} \\
    \includegraphics[width = 2.64in]{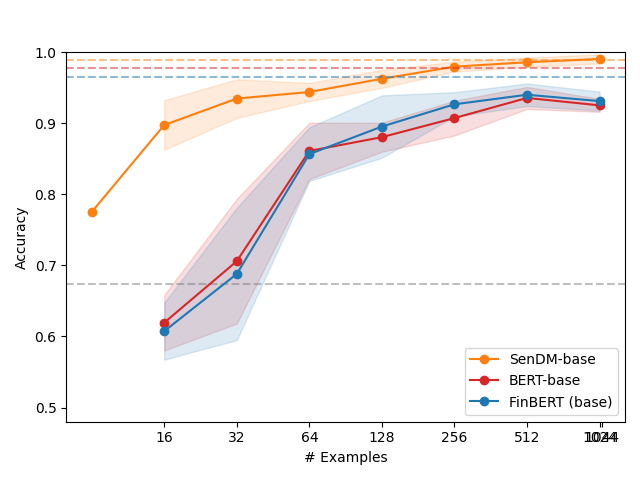}
 \end{tabular} &
  \begin{tabular}{c}
    tiny  \\ \includegraphics[width = 2.64in]{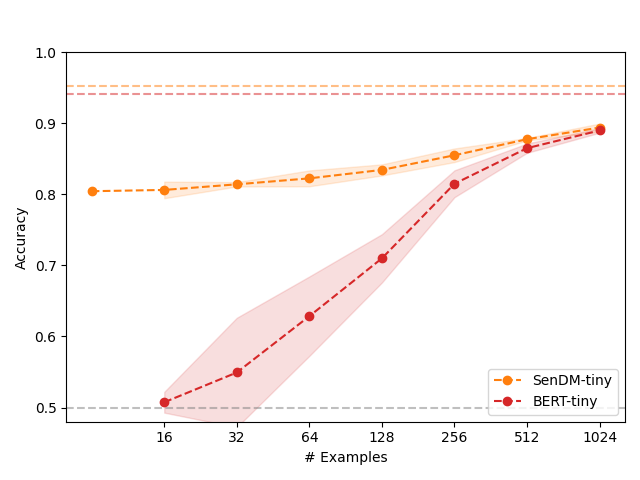} \\
    \includegraphics[width = 2.64in]{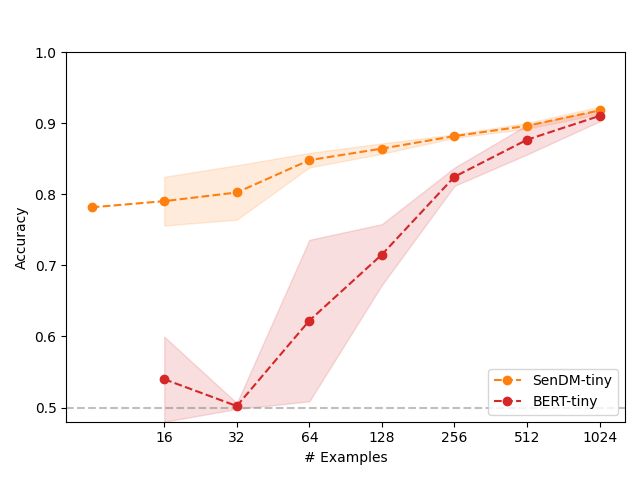}
   \\ \includegraphics[width = 2.64in]{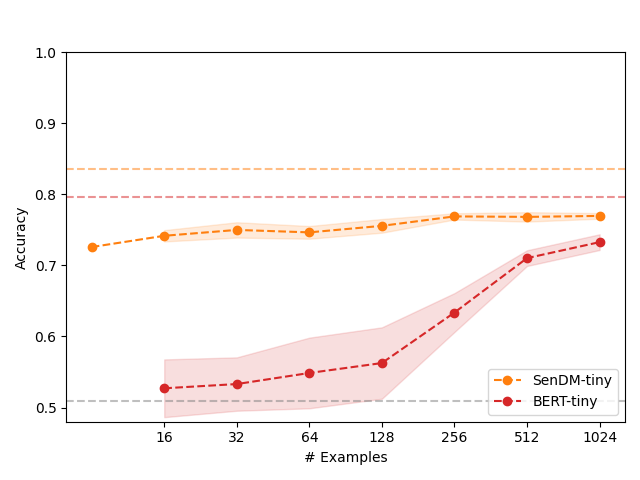} \\
    \includegraphics[width = 2.64in]{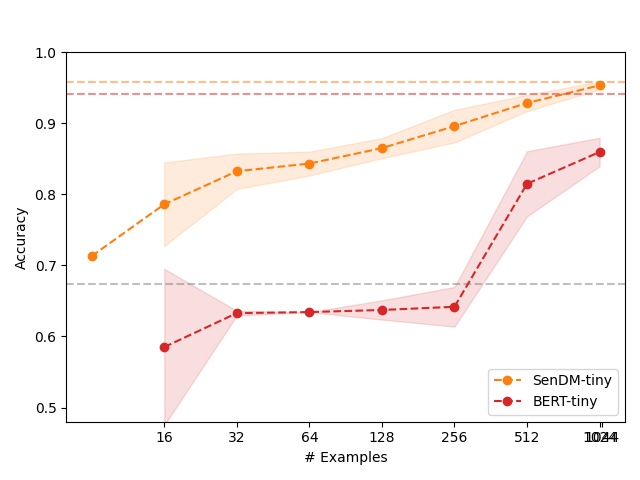}
    \end{tabular} 
\end{tabular}
\caption{Performance of \genmodel{} and baselines on four datasets given different amounts of training examples. Left column: base-size models. Right column: tiny-size models. Lines indicate the mean and shaded areas indicate the standard deviation over the five seeds (see section \ref{sec:eval} for details).  Dashed horizontal lines indicate the fine-tuning results for the full training data (the full-data setting). Dotted horizontal lines indicate the prior of the common class in the dataset. \finbert{} and \sentix{} are available only in the base size.}
\label{fig-general}
\end{figure*}
\endgroup

We now evaluate the performance of both the base and tiny versions of the general sentiment model (\genmodelbase{} and \genmodeltiny{}) on datasets from different domains. 
Since our main focus is on the zero and few-shot setting, 
we report 
the results after fine tuning over 
$0$ up to $1024$ training examples.

Figure \ref{fig-general} shows the accuracy of \genmodelbase{} (left column) and \genmodeltiny{} (right column), for all datasets, vs. the number of examples used for fine tuning. 
The accuracy is compared to that of vanilla BERT -- base and tiny respectively, and to \sentix{} \cite{zhou-etal-2020-sentix}, 
that are fine tuned over the same labeled examples. 
For \fbpds{}, 
we add the corresponding domain specific version of \bertbase{} -- \finbert{} \cite{Araci2019FinBERTFS}. 

In all datasets, \genmodel{} significantly outperforms the BERT baselines, including the finance-specific \finbert{}, especially when the number of examples used for fine tuning is 
relatively 
small.
The gain in performance is even more significant when focusing on the tiny architecture. 
This is especially evident in the \fbpds{} dataset, where \berttiny{} completely fails to learn with up to $256$ examples, whereas the \genmodeltiny{} is able to learn with as few as $16$ examples. A similar trend can be seen for the \sstds{} dataset.
As expected, the gap between \genmodel{} and its counterparts decreases with the increasing number of training examples, reflecting the decaying effect of the initial weights on the fine-tuned model. In most datasets, this gap completely vanishes in the full-data scenario, with the exception of \fbpds{}, where the full training-data is of size $1044$, only slightly larger than $1024$ which is the upper limit of our few-shot regime. 

From the stability perspective, \genmodel{} is more robust to changes in the initial seed compared to the other models, due to the lack of randomness in the initialization of its classification layer.  

\sentix{} is a sentiment-aware pre-trained language model that was originally designed for cross-domain review sentiment analysis. Importantly, \sentix{} is trained on large amounts of Yelp and Amazon reviews, along with their associated star rating, the same star rating used to define the training set and the test set in our
\amazonds{} and \yelpds{} datasets. Thus, one can not report zero/few-shot training results for this model in these two datasets, since the available model is already trained on large amounts of the respective train data. That said, it is intriguing to explore the performance of this model over our two other datasets. When considering the results in \sstds{}, which is based on movie reviews, we see strong performance for \sentix{}. This is expected, since this kind of data, composed of starred reviews -- albeit from a different domain -- is precisely the forte of \sentix{}. Interestingly, though, its gap compared to our \genmodel{} is relatively small, and insignificant when fine tuning over $16$ and $32$ examples. Considering the results in a more distant domain, namely the \fbpds{} dataset, where starred reviews are irrelevant, we see the clear value of our approach, that consistently outperforms all other models, including \sentix{}, typically by a significant margin, especially when labeled data is scarce. These results support our hypothesis that pretraining based on sentiment-related DMs will result in a more robust model, that yields superior performance when tested on various domains. 

A concern may arise, that the strong performance demonstrated by our approach on 
\fbpds  
is related to the fact that the general corpus giving rise to the weak labels used for inter-training, also contains some financial documents, and that the results will be inferior for domains not covered in the corpus we start with. 
To address this concern, we generate a version of \genmodel{}, in which 
financial documents are removed from the general corpus
\footnote{Based on topic tagging, see details in section \ref{sec:fin-exp-setup}.}. We find that there is no deterioration of results, supporting the notion that the 
observed 
improvement over alternative methods 
is not due to inter-training using financial documents -- see Figure 1 in the Appendix. 

Another concern we wanted to examine is related to the relevance of our approach for low resource languages, where a very large corpus like \gencorpus{} is not available. To this end, we checked the sensitivity of the results to the size of the weakly-labeled data, by creating two versions of \gencorpus{}, one based on inter-training using only $10\%$, and the other based on only $1\%$, of the weakly labeled data. Surprisingly, these two models resulted in no detrimental effect on the results.  In addition, one may also leverage the large English weakly labeled data for inter-training the multilingual BERT model (M-BERT). We leave the examination of this approach for future research.

To summarize, overall, the proposed 
DM-based sentiment model, significantly 
improves sentiment classification performance for both small and large language models. 
Remarkably, even the tiny version of \genmodel{} outperforms the much larger \bertbase{} baseline.

\section{Adapting \genmodel{} to a New Domain}
\label{sec:finance-model}
In section \ref{sec:resgenmodel} we saw that \genmodel{}, which leverages a general list of sentiment-related DMs, improves results over baselines on all datasets, including the finance dataset, \fbpds{}.
Here we investigate whether adapting \genmodel{} to a new domain, can further improve 
its performance on that domain. 
We choose to study this on the financial domain 
since as stated in \citet{Araci2019FinBERTFS}, financial sentiment analysis is a challenging task due to the specialized language and lack of 
domain-specific 
labeled data. 
Moreover, it is an important task for many potential users, and finally, the adaptation impact 
can be tested given the availability of the \fbpds{} dataset.

\subsection{The Training Approach}
\label{sec:dssentmodels}
The robustness 
of \genmodel{} 
presumably  
emerges from the multi-domain corpus it relies on, and the general nature of the DM list, $L_{g}$, 
composed of discourse markers that are abundantly used and carry 
a general sentiment signal. 
However, due to potential domain-specific jargon and language style, given a domain specific text corpus $C_d$, it may be useful to build domain specific sentiment models.

We study five ways to build domain specific sentiment models based on the 
general 
flow described in Figure \ref{fig:sketch}. 
All five 
resulting 
models, described in the bottom part of Table \ref{tab:models}, are based on weakly labeled sentences from the domain-specific corpus, $C_d$.
All five models rely on the availability of a general sentiment 
model, trained in a domain-independent manner, such as \genmodel{}, which we release to the community. 
In the experiments we describe here, we use a variant of \genmodel{}, denoted below \genmodelnonfin{}, which is developed as \genmodel{}, but after removing finance-related documents from the general corpus $C_g$, to better simulate the finance domain as a new domain.\footnote{In practical applications, though, users can obviously use our \genmodel{} directly when building a domain-specific model.} This \genmodelnonfin{} model is used as the straining point for inter-training all five models, and in some cases to define the inter-training weakly labeled data, as described next. 

The first model, \dmodelLg{}, uses the general DM list, $L_g$, for weak label extraction
from text in the target domain. 
However, since 
sentiment-related DMs might be domain specific 
we develop a method 
to extract a domain-specific sentiment-related DM list, $L_d$.
To that end, we note that 
there is no need to define a DM using the standard linguistic definition, 
and a functional definition can be used instead. 
Thus, we 
define as a sentiment-related DM , any n-gram ($n<=3$)
followed by a comma, 
for which the set of sentences it opens is enriched with highly confident positive/negative predictions, as determined by \genmodelnonfin{}. The second model, \dmodelLd{}, relies on a list composed of such DMs instead of $L_g$.
As a third approach, we 
perform one step self-training, 
where the high confidence predictions of \genmodelnonfin{}  
over sentences from $C_d$ are used for inter-training, ignoring the DMs.\footnote{In this work we use predictions with $score>0.9$ and $score<0.1$ as positive/negative weak labels.} 
This model is denoted by \dmodelp{}. 
Finally, aiming to reduce labeling noise, we explore 
a synergistic approach, i.e., taking as 
weakly labeled 
positive/negative sentences only those for which 
an 
opening DM conveys a sentiment signal, 
and that sentiment is consistent with the high confidence prediction of 
\genmodelnonfin{}. 
We study this approach with our two DM lists, $L_{g}$ and $L_{d}$, giving rise to 
two additional models, denoted \dmodelLgP{} and \dmodelLdP{}, respectively. 

\begin{table}[ht!]
\begin{center}
\renewcommand{\arraystretch}{1.6}
\begin{tabular}{p{2cm}| p{0.6cm} |p{0.6cm} |p{1.3cm} | p{1.5cm}} 
 \hline
 \textbf{Model name}
 & \textbf{C} & \textbf{L} & \textbf{M} & \textbf{With self-training} \\ 
 \hline
 \genmodel & $C_g$ & $L_g$ & BERT & NA\\ 
 \hline\hline
\dmodelLg & $C_d$ & $L_g$  & \genmodel & No\\
 \hline
 \dmodelLd & $C_d$ & $L_d$ & \genmodel & No\\
 \hline
 \dmodelp & $C_d$ & NA  & \genmodel & Yes\\
 \hline
\dmodelLgP & $C_d$ & $L_g$  & \genmodel & Yes\\
 \hline
 \dmodelLdP & $C_d$ & $L_d$ & \genmodel & Yes\\
 \hline
\end{tabular}
\end{center}
\caption{Sentiment language models and the corresponding assignment of $C$, $L$ and $M$ in the flow described in figure \ref{fig:sketch}, as well as whether the predictions of \genmodel{} are used for assigning weak labels ("With self-training" -- see main text for details) . \genmodel{} is a general (multi-domain) model. The other five are domain specific. $C_g$: a general, multi-domain, corpus; $C_d$: a corpus from domain $d$; $L_g$: a list of DMs associated with sentiment in the English language; $L_d$: such a list adapted to domain $d$.
\label{tab:models}
}
\end{table}

\subsubsection{Domain Specific Sentiment-Related DMs}
We now turn to describe how we generate the domain specific sentiment-related DM list, $L_d$, given the domain specific corpus $C_d$ and using \genmodelnonfin{}. 
Note, 
we are not interested in identifying all domain specific sentiment related DMs. Rather, we are seeking a precision oriented list, for the purpose of obtaining a 
high-quality 
weakly labeled set of sentences for 
inter-training process. Thus, we perform strict automatic filtering in the process.  
The idea is to first identify all n-grams that may potentially serve as DMs, and then identify whether the set of sentences they open is enriched with positive/negative sentiment, based on the predictions of \genmodelnonfin{}. 

The first step consists of \textbf{identifying a list of candidate DMs}. To this end we first identify all unigrams, bigrams, and trigrams, that, followed by a comma, open sentences in $C_d$, and further group these using NER (e.g., instead of multiple bigrams of the type "on Sept 9th", "on 10/2/2020",... we generate one bigram "on DATE"). The candidate list consists of the $1000$ most frequent DMs. Specific filters may be further applied depending on the domain corpora -- see appendix for details.

The second step consists of \textbf{using \genmodelnonfinbold{} to select the domain-specific DMs out of the candidate list}. 
We analyse the sentences that start with the DMs in the above candidate list, to find those DMs whose 
associated 
sentences are significantly associated with a 
highly confident prediction of 
positive/negative sentiment. 
For each candidate DM, we 
sample $1000$ sentences from the set of all sentences that start with the DM followed by a comma, and assign 
each of these sentences with 
a sentiment if it is scored with high confidence\footnote{$<0.1$ or $>0.9$} by \genmodelnonfin{}. 
For each candidate DM, we perform a statistical analysis of the sentiment of its 
associated 
sentences, on the sample of $1000$ sentences, provided that they are not too repetitive based on token counts, and that the sentiment class with the higher count comprises at least $85\%$ of the sentences assigned a sentiment. 
A DM is considered to be associated with a positive/negative sentiment if the p-value of the positive/negative class is smaller than $0.01$ after Bonferroni correction for multiple tests, based on a Hypergeometric test.
We release the code that allows a user that has a  corpus of interest to use \genmodel{} to generate a specific DM list adapted to the corpus, as described above.

\subsection{Experimental Setup}
\label{sec:fin-exp-setup}
The inter-training and evaluation details are identical to those described in Section \ref{sec:gen-exp-setup}.
For the number of weakly labeled samples used for inter-training the finance-specific sentiment models, see Table 1 in the appendix. 
In all cases we divide the samples into training ($80\%$), development ($10\%$), and test ($10\%$) sets. 

\subsubsection{Finance-specific Corpora}
From the General Corpus, $C_g$, we can define a sub-corpus that is focused on the financial domain, using a provided metadata topic field, and filtering only for articles from the topic 'Finance'. We term this corpus $C_{fin}$ and use it for studying adaptation to the finance domain.

\subsubsection{\genmodelnonfinbold{}.} As we are interested in studying the scenario of adapting to a new domain, which is possibly not covered by the data used to train \genmodel{}, and since the corpus used to train \genmodel{} does contain some financial documents, we generate a variant of \genmodel{} 
excluding the finance domain. 
This model, denoted by \genmodelnonfin{}, is trained analogously to \genmodel{}, except we exclude financial documents from \gencorpus{} before using it. 
Naturally, we do not expect a user to train such a model, it is used here and in the appendix, only to 
examine to what extent our approach can generalize
to domains not covered by the general corpus used to train \genmodel{}.

\subsection{Results}

\begin{figure}
\includegraphics[scale=0.55]{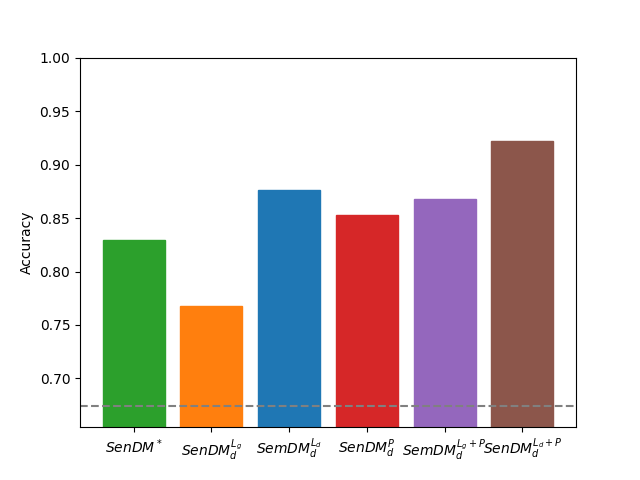}
\caption{Performance of the general model and the various domain specific models on the finance dataset, \fbpds{}, for the zero shot setting. $\ast$: \genmodelnonfin{} is the general model when trained on a corpus not containing financial documents -- see main text for details. All models are base size. In all models the domain $d$ is finance. The dashed horizontal line indicates the prior of the common class in \fbpds{}.
\label{fig:finzeroshot}
}
\end{figure}

Leveraging \genmodelnonfin{} and $L_{fin}$, we generate the five versions of domain specific sentiment models described in \ref{sec:dssentmodels} (see Table \ref{tab:domaindms} for the DMs in $L_{fin}$\footnote{We also use \genmodelnonfin{} instead of \genmodel{} to identify $L_{fin}$ for the reasons described above}.).
Figure \ref{fig:finzeroshot} depicts the accuracy of the finance specific sentiment models, on the finance dataset \fbpds{}, for the zero shot setting, in comparison to the accuracy of the general model, \genmodelnonfin{}.
As can be seen, indeed using the domain specific DMs rather than the general list improves the accuracy (blue vs. orange, and brown vs. purple bars).
One step self-training is valuable even without combining it with a signal from DMs (red vs. green bars).
Combining one step self-training in a synergistic manner with the signals from the DMs as described above brings additional value (purple vs. orange and brown vs. blue bars).
The highest accuracy is achieved when combining the signal of the finance-specific DMs with one step self training (brown bar).
All above accuracy comparisons are significant ($p<0.05$), based on McNemar’s test.
It is interesting to note that using $L_g$ for extracting the weak labels from the finance corpus, results in lower performance than using the general model (orange vs. green bars). 
This may be attributed to a higher noise in the signal of the $L_g$ DMs in the finance corpus compared to the general corpus. 
This explanation is consistent with the improvement gained by incorporating  self-training in a synergistic manner with $L_g$, a step that results in noise reduction. 
For the few shot scenario we do not find significant differences between the models -- see Figure 2 
in the Appendix.

We note that the 
suggested approach may be applied iteratively to gain further improvement on the finance dataset. Moreover, this adaptation process can also be applied to \genmodel itself. We leave these directions for future work.

\subsubsection{Analysis of Domain Specific Sentiment DMs}
As we saw above, leveraging the finance specific DMs, was useful for adapting the sentiment model to the finance domain. 
Table \ref{tab:domaindms} lists the sentiment-related DMs extracted from general English, $L_g$, as well as from the financial domain. 

\begin{table}[ht!]
  \centering
\begin{tabular}{p{1cm}| p{3.1cm} | p{3.1cm} } 
 \hline
 \textbf{Domain}
 &  \textbf{Associated with Positive Sentiment} &  \textbf{Associated with Negative Sentiment} \\ 
 \hline
 General 
 & 'fortunately', 'happily', 'hopefully', 'ideally', 'luckily', 'thankfully' & 'admittedly', 'curiously', 'inevitably', 'sadly', 'unfortunately' \\
 \hline\hline
 Finance & 'as ORG', 'at the event', 'fortunately', 'hopefully', 'ideally', 'in future', 'in other business', 'luckily', 'once completed','ORG CEO', 'starting DATE', 'thankfully', 'the program', 'this way', 'to achieve this', 'under his leadership', 'with ORG' & according to police', 'sadly', 'the problem', 'the problem is', 'unfortunately', 'worse' \\
 \hline
Sports & 'beginning DATE', 'fortunately', 'in the future', 'luckily', 'thankfully', 'that way' & 'admittedly', 'alas', 'granted', 'ironically', 'sadly', 'true', 'unfortunately', 'unfortunately for ORG' \\ 
 \hline
         Science & 'established in DATE', 'if necessary', 'if possible', 'if successful', 'luckily', 'that way', 'to address this', 'when possible', 'whenever possible', 'where possible', 'with this approach' & 'admittedly', 'at ORDINAL glance', 'at times', 'curiously', 'even then', 'even worse', 'in part', 'inevitably', 'paradoxically', 'predictably', 'regrettably', 'the problem', 'there was', 'too often', 'unsurprisingly', 'without it', 'women'\\
 \hline
\end{tabular}
\caption{Sentiment-related DMs. The lists below the double line are domain specific. The upper case tokens are NER tags.
\label{tab:domaindms}
}
\end{table}
Although some of the finance-specific DMs echo those appearing in general English 
usage 
(e.g., 'fortunately'), many DMs are domain specific, and in fact may not be considered a DM by the standard linguistic definition. 
For example, the bigram 'ORG CEO' ('Walmart's CEO', 'BOA's CEO', etc.), is associated with a positive sentiment. 
This might 
be surprising at first sight, and would probably not be listed in a manually curated finance-specific DM list, but in hindsight makes sense. When considering what sentences might follow such an opening, one would expect them to discuss positive things about the company. Another such example is 'under the leadership'. Here again, we find, that although not expected 
{\it a-priori\/}, most sentences that follow this opening, would carry a positive sentiment due to the reference to leadership.

Next, we were interested to see how sentiment DMs vary in other domains. 
We carved out several domain corpora out of the general corpus. Beyond the \fincorpus{} described above we also introduce: (1) \textbf{The \sportscorpus{}}: Similarly to the creation of the \fincorpus{} but filtering for articles from the topic 'Sports'; and (2) \textbf{The \scicorpus{}}: This corpus focuses on scientific articles from scientific journals, and is defined as any article in \gencorpus{} that was published in one of the journals included in the list of journal impact factors\footnote{ https://www.scimagojr.com/journalrank.php}. 

These lists can be found in Table \ref{tab:domaindms}. We find that some DMs continue to be ubiquitous across domains (e.g., 'fortunately'), but others seem to be associated with sentiment in specific domains only. 
An interesting example is the word 'women', which in the scientific domain, is among the list of negatively associated DMs. We found that in scientific papers, sentences beginning with 'women' will often deal with issues like oppression and deprivation, and thus are associated with a negative sentiment. 



\section{Discussion}


This work suggests to leverage DMs that carry a sentiment signal to inter-train and adapt general-purpose language models to the sentiment classification task. The obtained sentiment analysis models demonstrate significant performance boost across multiple domains, most notably in the zero-shot and few-shot learning scenarios, emphasizing the practical value of this work. We further show how to evolve the obtained models to a specific domain of interest using automatically identified domain-specific DMs, and show how this approach yields a further performance enhancement in zero-shot learning within the challenging finance domain. 

The ability to bootstrap a general, small, and easily identified seed of sentiment carrying DMs into a powerful sentiment analysis model may hold additional valuable implications. For example, this approach can be easily adapted to languages beyond English, including low resource languages, as long as a reasonably sized corpus is available. Another interesting direction would be to expand
the proposed approach for {\it targeted\/} sentiment analysis. For example, in the finance domain, the company appearing in a sentence can be considered as the sentiment target.

\citet{sileo2020discsense} show that various NLP task classes are naturally associated with specific DMs. Thus, the methodology presented here for leveraging DMs to create task-specific language models can be potentially applied to tasks beyond sentiment analysis.  
 Finally, DMs probably represent only one type of linguistic cues among the richness of signals in natural language that can be leveraged as self-supervision to align LMs with downstream tasks.

\appendix

\section{Appendix}
\label{sec:appendix}

\subsection{Filters Applied to the Domain-Specific DM Selection}
In some domains certain DMs may be very specific to a journal, or even a reporter. We find this is the case in the finance domain. Hence, we further filter out DMs whose sentences originate from a relatively narrow set of journals. For this purpose we define the entropy of a DM, based on the probability distribution of its sentences across journals, and filter out $30\%$ DMs with lowest entropy. 
Furthermore, in the case of the finance domain, since the sentiment analysis task is to identify a sentiment with respect to a company, we apply the above process only to sentences mentioning a company name, where we use the 
list of all companies traded in one of the five major stock exchanges\footnote{Note that this sentence selection step is not applied to the finance corpus which contains all sentences from finance documents}.

\subsection{Additional Tables and Figures}
\begin{table}[ht!]
\begin{center}
\renewcommand{\arraystretch}{1.6}
\begin{tabular}{p{2.5cm}| p{4.5cm} } 
 \hline
 \textbf{Model name}
 & \textbf{Total number of samples used for intermediate training} \\ 
 \hline
 \genmodel & 1,876,614\\ 
 \hline
 \genmodelnonfin & 1,815,943\\ 
 \hline\hline
\dmodelLg & 60,671\\
 \hline
 \dmodelLd & 99,521\\
 \hline
 \dmodelp & 490,989\\
 \hline
\dmodelLgP & 45,246\\
 \hline
 \dmodelLdP & 70,681\\
 \hline
\end{tabular}
\end{center}
\caption{The number of samples used as weak labels to train each sentiment model. In all cases we divide these into training ($80\%$), development ($10\%$) and test ($10\%$) sets.
\label{sup-tab:sizes}
}
\end{table}

\begin{figure}
\includegraphics[scale=0.5]{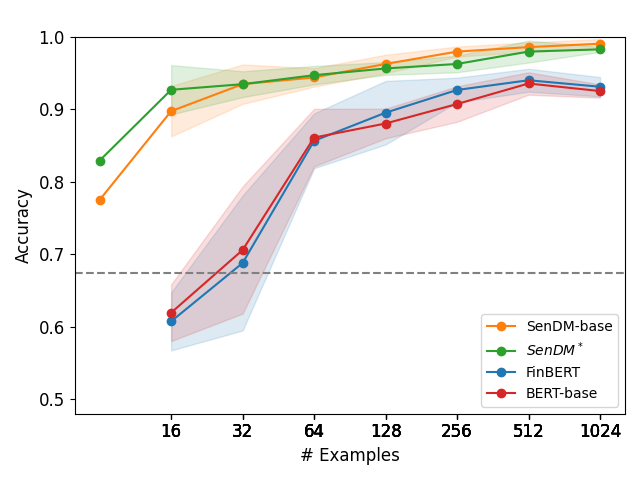}

\caption{Performance of \genmodelbase{} and base sized baselines on \fbpds{} given different amounts of fine-tuning examples. Lines indicate the mean and shaded areas indicate the standard deviation over the five seeds (see Section 3.4 for details). The dashed horizontal line indicates the prior of the common class in the dataset. 
\genmodelnonfin{} is the same as \genmodel{} when  trained excluding financial documents.}
\label{app-fig:fin}
\end{figure}

\begin{figure}
\includegraphics[scale=0.5]{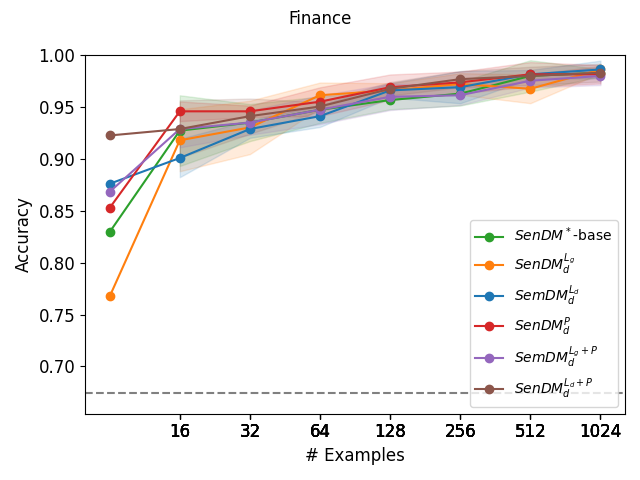}
\caption{Performance of the general model and the various domain specific models on the finance dataset \fbpds{} for the zero and few shot settings. $\ast$: \genmodelnonfin{} is the general model when trained on the general corpus excluding financial documents -- see Section 4.2 for details. In all models the domain $d$ is finance. The dashed horizontal line indicates the prior of the common class in \fbpds{}.}
\label{fig:finresultsfull}
\end{figure}

\bibliography{main}

\end{document}